# From Tool Use to Technological Agency: LoopCAT as a Local-First, Open-Source Tool for Translation Technology Education

Gokhan Dogru — Universitat Pompeu Fabra — gokhan.dogru@upf.edu

Adrià Martín Mor — Universitat Autònoma de Barcelona — Adria.Martin@uab.cat

## Abstract

Translation students need to learn both how to use translation technologies and how to judge the choices those technologies make available. This article presents LoopCAT, an Apache-2.0-licensed, local-first computer-assisted translation environment co-created with OpenAI Codex using GPT-5.5 and GPT-5.6, and proposes a framework connecting workflow competence, evaluative judgement, and technological agency. The account draws on repository history, implementation inspection, and the verification records of an identified development build. LoopCAT combines local project storage, translation memories, terminology, quality assurance, document exchange, and optional connections to local or hosted AI services. Its English, Catalan, and Turkish interface catalogs also make the application itself available as teaching material: students can translate English UI strings into another language, review the existing automatically generated target drafts, import their revisions, and test the interface. We organize these opportunities around four forms of participation: operating a workflow, evaluating outputs, inspecting and configuring mechanisms, and making or defending a bounded intervention. A six-session sequence, a UI-localization assignment, a placeholder example, and an assessment rubric specify how teachers could use the framework. The paper separates implemented capabilities from proposed educational benefits; it reports no new student-learning outcomes. It distinguishes the latest package checks from earlier regression evidence and sets out a protocol for classroom evaluation. LoopCAT provides an inspectable setting for teaching how translation decisions interact with data, interfaces, and software rules. Whether these activities improve judgement, transfer, or participation remains an empirical question.

Keywords: translator education; computer-assisted translation; technological agency; evaluative judgement; AI literacy; local-first software; open-source pedagogy; translation data; software localization.

## 1. Introduction

Learning a translation tool and learning to make decisions about translation technology are related but different educational objectives. A student may know how to import a document, accept a translation-memory match, or request an AI suggestion without being able to explain why the suggestion is appropriate, where the source text has been sent, or whether the resulting bilingual data can be reused elsewhere. Conversely, a student who can discuss AI ethics in general terms may struggle to identify the practical consequences of a provider setting or an export option. Translation technology education needs opportunities to connect these forms of knowledge within actual workflows.

Fluent machine-generated language can conceal semantic, terminological, or structural problems. Students need to establish quality criteria, examine evidence, compare alternatives, and take responsibility for a decision. They also need to examine how translation environments organize work: which suggestions appear, what information remains visible, what counts as an error, and which actions require connectivity or an external account. A translation technology course can investigate these arrangements alongside the translated text.

The educational question motivating LoopCAT is therefore: how can a translation environment support learning about both the production of translations and the technological conditions of that production? The project approaches this question through a local-first, open-source computer-assisted translation (CAT) application. Its immediate engineering objective is a dependable editing and delivery workflow. Its pedagogical objective is to make that workflow a site for observation, explanation, experimentation, and justified intervention. The second objective depends on the first: students cannot readily investigate translation decisions if they are primarily occupied with recovering lost work or diagnosing an installation.

This article examines a functioning software artifact and its possible uses in teaching. The software was co-created through collaboration between the developer and OpenAI Codex, which contributed to design, implementation, testing, and documentation. Section 4.4 examines that process as part of the project's account of technological agency. It explains how selected architectural and interface decisions support activities in which students can inspect and change parts of a translation workflow. It then specifies a teaching sequence, examples, and assessment evidence. Finally, it identifies the classroom research needed to test those proposals. The claim is about opportunities the implementation makes available, not demonstrated learning gains or a requirement that translators become programmers.

The intended audience includes translation educators, researchers in translation technology, and developers of educational language tools. LoopCAT can complement commercial CAT instruction, existing open-source tools, and dedicated machine-translation teaching platforms. A public release is planned for 1 September 2026. This revised account identifies the 31 August development preview and distinguishes the checks repeated for its documentation and packaging refresh from the broader verification of the preceding build; it does not report a completed public release.

## 2. Related work and conceptual foundations

### 2.1. From operational competence to informed judgement

The European Master's in Translation competence framework places technology alongside translation, language and culture, personal and interpersonal competence, and service provision (EMT Board and Competence Task-Force, 2022). This positioning matters: technical operation is part of professional competence, but does not replace linguistic judgement, ethical responsibility, or communication with other participants in a workflow. Although the framework addresses master's programmes, its integration of these dimensions can also inform appropriately scaled undergraduate activities; using it here does not imply EMT accreditation or equivalence between degree levels.

Tai et al. (2018) conceptualize evaluative judgement in terms of students' capacity to determine the quality of their own work and that of others. Their discussion of exemplars, assessment criteria, self-assessment, peer assessment, and feedback provides a useful basis for translation teaching. A correct final answer alone reveals little about whether students can recognize quality independently. The educational task must also make their criteria and reasoning available for examination.

Bearman et al. (2024) extend this discussion to generative AI, distinguishing the evaluation of AI-produced work, the evaluation of AI processes, and the use of AI to assess students' own evaluative judgements. Applied to translation, this perspective encourages teachers to ask why a candidate was accepted, what the prompt or workflow made available, and whether an automated assessment deserves confidence. LoopCAT's proposed pedagogical use connects such evaluative activity to persistent resources and constraints: a candidate belongs to a project with terminology, context, previous translations, and delivery requirements.

Dogru's (2026) classroom case study provides a relevant, but distinct, empirical point of departure. It examines 23 student projects, including 22 written reports, from a machine-translation and post-editing course. Students compared AI-mediated outputs, used automatic and human evaluation, and justified post-editing choices. The study documents ways in which students negotiated different forms of evidence; it is not a controlled test of LoopCAT and does not establish that a particular tool caused learning gains. The present article extends the educational question from judging outputs to examining and, where appropriate, changing the workflow that produces and presents them. It introduces no new analysis of those students' data.

### 2.2. AI literacy, translation data, and the legitimacy of refusal

Long and Magerko's (2020) account of AI literacy links effective interaction with critical evaluation and an understanding of AI's capabilities and limitations. Translation-specific work makes the data dimension particularly visible. Hackenbuchner and Krüger (2023), through DataLitMT, connect professional machine-translation literacy with the ability to work critically with data. Their basic and advanced learning resources provide a precedent for teaching data inspection through tasks matched to learners' prior knowledge.

Two earlier studies by Dogru provide concrete connections to this argument. Dogru and Moorkens (2024) investigate translation-memory data augmentation for desktop NMT fine-tuning across three language pairs. Dogru (2024) describes the construction of domain-specific translation memories for fine-tuning. These studies bring attention to corpus selection, preparation, and evaluation as forms of translator participation in technological production. They do not establish that every data intervention improves a model, and LoopCAT should not be represented as implementing their complete training pipelines. Their relevance is the broader principle that translation expertise can inform decisions about resources and systems, as well as individual target segments.

Ethical participation includes deciding not to use a technology. Moorkens and Doğru (2026) propose teaching activities that connect AI ethics to practical cases, stakeholder relationships, and the consequences of action. In the present framework, declining to transmit a confidential text, rejecting an unsuitable suggestion, or choosing an adequate non-AI workflow can demonstrate agency. Assessment should not reward the mere frequency of AI use or assume that accepting automation is the desired outcome. Students should be able to defend both adoption and refusal in relation to a brief, evidence, and the people affected.

### 2.3. Open translation technologies as educational precedents

LoopCAT builds on an established tradition of open translation technology. OmegaT provides a free translation-memory environment with terminology resources and invites contributions through interface localization, documentation, and testing (OmegaT project, n.d.). MTradumàtica connects machine-translation customization with a modular teaching proposal (Martín-Mor and Piqué, 2017). MutNMT introduces non-specialists to NMT through a web interface; its expert role permits model training within limits on time, data, and parameters (Ramírez Sánchez, 2023). Kenny's (2022) edited open-access volume brings together approaches to teaching machine translation and supporting informed use.

These precedents establish that open CAT software, accessible MT education, and translator contributions to software are not new. LoopCAT brings a local CAT workflow, optional AI connections, and editable interface resources into one application. This paper proposes a teaching sequence that connects those resources to inspection and intervention. It does not compare learning outcomes with OmegaT, commercial products, or specialized MT platforms.

OPUS-CAT offers a further architectural precedent: desktop neural machine translation integrated with CAT environments and local fine-tuning (Nieminen, 2021). It should be distinguished from a large language model. LoopCAT can connect to OPUS-CAT as an NMT service; it can also connect to local LLM-serving applications such as Ollama and LM Studio. These are different components with different functions and licensing conditions. Making these distinctions explicit is itself an educational objective.

### 2.4. Local-first design and a bounded definition of agency

Kleppmann et al. (2019) describe local-first software through ideals that extend beyond working without a network: responsiveness, user ownership, longevity, privacy, and collaboration across devices also matter. LoopCAT currently implements a subset of this orientation through local project data, an offline-capable editing environment, and portable exports. It does not implement the full collaborative and synchronization model discussed in that work. The term local-first is used here to identify a design direction and concrete data-handling choices, not to claim conformance with every ideal.

For this article, technological agency means a learner's situated capacity to explain, select, configure, challenge, and, where appropriate, alter the technological conditions of translation work. This is a working definition proposed for the present design framework, not a validated psychological construct or measurement scale. Agency depends on knowledge and opportunities, but also on institutional rules, time, hardware, available support, and the permission to refuse an option. Source-code access creates a possibility; it does not ensure that a learner can exercise it.

The project also seeks greater technological sovereignty. Here that ambition is bounded to practical questions: who can access the data, which dependencies are required, whether work can continue without a service, and whether a workflow can be transferred or changed. A local application does not eliminate dependence on operating systems, processors, model providers, or developer maintenance. It can make some of these dependencies easier to identify and negotiate.

## 3. Materials, method, and scope of the account

This is an artifact-centered design analysis, not an intervention study. The materials comprise the LoopCAT repository, selected historical commits and tags, project documentation, implementation files, local verification records, and retained Codex development-task records. For the co-creation account, user-visible exchanges and tool-operation records establish examples of specification, execution, and correction; session metadata establishes recorded model and mode settings. These are selected examples, not a complete census of development activity. The developer's account supplies information, such as varied speed settings, not established by the inspected metadata. Current OpenAI documentation clarifies terminology rather than proving historical use. Repository history supports the chronology; source inspection supports the capability descriptions. These checks were supplemented by targeted execution of existing localization modules and the placeholder detector and QA rule used in the examples below. The verification records support claims about two identified 31 August builds, within the different scopes described below. Literature was selected to connect this work to translator education, evaluative judgement, AI and data literacy, and open educational technologies; this is not a systematic literature review.

The unit of analysis is a design decision or capability and its proposed educational consequence. For each, the analysis asks what students can do, which mechanism or resource they can inspect, what evidence an educator could assess, and which limitations interrupt the connection. The resulting mappings are interpretive design propositions. They should be tested through classroom research rather than read as observed student behaviour.

The packaged baseline is LoopCAT 0.0.4-dev.20260831, build identifier 0.0.4-dev.20260831+source.239cf868ce3d. Its 645 fingerprinted source files match committed source ef4f8c7e5e1d957915aaa4fb607b84709c5ee217; download-record commit 21ca84a0bea6791633dedb5c213f4e7203a14afa preserves that source identity. The full source fingerprint, artifact checksums, and check scope are recorded in downloads/release.json and downloads/verification.md at that download-record commit (LoopCAT project, 2026). This build supersedes the earlier 30 August package and the first 31 August build, 0.0.4-dev.20260831+source.92e3832c0146. The latest refresh omits internal review and planning documents from desktop packages and personal workspace references from tracked material. Its verification record distinguishes fresh package checks from the preceding build's full regression and installed-upgrade results. A separate audit for this revision checked the three ZIP hashes, embedded build identities, and correspondence with the committed source. Section 5.3 specifies what was and was not exercised; the planned public release must retain an equally explicit record.

Three levels of claim are maintained throughout. Implemented refers to a capability found in the inspected source, within the identified source snapshot. Verified refers to a capability or artifact exercised by the recorded checks within their stated environment. Proposed refers to a teaching activity, learning mechanism, or future evaluation that has not yet been established through a LoopCAT classroom study. The project's developer is involved in this account, creating a risk of favourable selection and interpretation. The proposed evaluation therefore includes counterexamples, independent assessment, and publication of the teaching materials.

## 4. Development history as a sequence of educationally relevant choices

### 4.1. Establishing a reliable translation loop

The earliest available repository commit, dated 6 July 2026, already contains a substantial application. It establishes a lower bound for the documented history, not the date on which the project was conceived. Early documentation centers on an offline translation workflow: importing material, working through segments, using resources, reviewing output, and completing delivery without losing work. These records establish a practical objective: keeping the editing and delivery workflow usable offline. They do not establish when the developer first formulated the teaching aims discussed here.

The July development record expands this foundation through localization of the interface, translation-memory and terminology work, quality and focus-oriented interaction, and document interchange. A 0.0.2 checkpoint appears on 9 July. OPUS-CAT integration and web-bridge work make local MT a possible part of the workflow, while XLIFF work strengthens exchange with other environments. In the inspected local repository, draft-0.0.3 resolves to commit 6f9754d, dated 16 July. Later features should not be retrospectively attributed to that historical tag.

For teaching, this phase brought familiar CAT concepts into a shared workflow. Segmentation, reuse, terminology, review, and export are not separate demonstrations: a choice in one part of the workflow can affect the others. A terminology decision may improve consistency while requiring revision of a memory match. A fluent target may fail because it damages a placeholder. A completed translation may still require verification in its exported format.

### 4.2. Making the environment more intelligible and maintainable

August work develops the application's identity and usability while also addressing internal complexity. The Loopbird branding introduced in early August expresses a local-first workspace organized around human control, visible feedback, and continuity of work. These are stated design commitments, rather than evidence that users already experience the software in that way.

The subsequent restructuring made particular responsibilities easier to locate in the source. The development record documents extraction of responsibilities into services and controllers, clearer build boundaries, and separation of test machinery from the production renderer. These changes can make a guided trace from an interface action to its implementation more feasible. They do not remove all complexity: a small entry file can still lead into a large composition layer. The appropriate classroom task is consequently a bounded investigation of a rule or pathway, not an expectation that beginners understand the entire application.

Interaction work also illustrates how user control becomes a concrete engineering concern. The August keyboard pass prioritizes browser-safe shortcuts and avoids interference with text input. The command palette is available through F2 or Ctrl/Cmd+Shift+P. Quick Insert uses existing candidates rather than silently initiating a new AI request. Such details matter because an apparently convenient interaction can otherwise create an unexpected network action or disrupt multilingual typing. They offer useful cases for discussing whose assumptions are embedded in interface design.

### 4.3. Treating release identity as part of reproducibility

By 30 August, a further problem was the relationship between the substantially modernized source and downloads still associated with older 0.0.3 labels. Release preparation introduced a distinct development version, rebuilt web and Windows artifacts from a shared source snapshot, and recorded their identities and checksums. Verification examined the actual packaged contents, rather than inferring their identity from filenames. Historical releases were kept distinct from the then-current development build.

On 31 August, interface refinements and renewed verification produced a new development preview. A further documentation and packaging refresh removed internal review and planning documents from the desktop payload and sanitized personal workspace references in tracked material. That refresh produced the current build identified in Section 3. Its recorded comparison found 37 renderer/runtime files byte-identical to the preceding verified build, but the full regression suites were not repeated. This distinction prevents a narrower packaging check from being reported as a fresh run of the entire test programme.

Teachers and researchers need an identifiable software version. Teachers need to know that a worksheet, a shortcut, and an installed application refer to the same behaviour. Researchers need to identify what participants used. Students comparing results need to distinguish a changed model or application from a changed translation decision. Versioning and provenance therefore belong within methodological literacy, rather than being treated solely as distribution administration.

Table 1. Selected milestones and their interpretation for teaching.

| Period | Documented development | Educational interpretation and boundary |
|---|---|---|
| 6–9 July | Existing editing workflow; open-source licensing; interface localization; TM/TB development; 0.0.2 checkpoint | A foundation for workflow practice. The available history does not establish the original conception date. |
| 9–16 July | Local MT connection and bridge work; XLIFF development; draft-0.0.3 tag | Opportunities to investigate interoperability and local MT. Later August features are outside this tag. |
| 7–24 August | New identity and interface work; incremental restructuring; storage, recovery, export, and desktop hardening | Conditions for more inspectable and dependable exercises. Architectural improvement is not evidence of learning gains. |

| Period | Documented development | Educational interpretation and boundary |
|---|---|---|
| 25 August | Browser-safe keyboard and candidate-insertion refinements | A concrete case in multilingual input and intentional AI use. |
| 30 August | Distinct 0.0.4 development identity; rebuilt downloads; source and artifact verification | An earlier identifiable baseline, subsequently superseded. Historical releases remain distinct. |
| 31 August | Updated preview, broader verification, then a documentation and packaging refresh with a new source identity | The current baseline separates fresh package checks from earlier regression evidence. Public release and remaining qualification still require explicit records. |

Source: repository history, release notes, and local verification records (LoopCAT project, 2026). Separate experimental branches and the distinct LoopCAT Plus exploration are excluded from the capability claims in this article.

### 4.4. Human–AI co-creation with OpenAI Codex

LoopCAT was co-created by the authors working with OpenAI Codex. Codex contributed to software design, implementation, refactoring, testing, and documentation, well beyond language editing of this article. The developer supplied translation-workflow requirements, selected priorities, evaluated behaviour, reported problems, and authorized integration of changes. Retained task metadata confirms use of GPT-5.5 and GPT-5.6 Sol with different reasoning-effort settings. The developer also reports varying speed settings. Model choice, reasoning effort, and service tier are separate controls; the inspected metadata does not establish the speed tier used for each task (OpenAI, n.d.-b). No comparison of model performance or proportion of AI-written code is claimed.

Planning and persistent goals. Planning was used to separate decisions about scope from implementation. The retained modernization prompt required repository inspection, preservation of offline operation and data compatibility, reviewable work packages, and explicit verification criteria. Recorded Plan-mode use supports this account; a written plan alone would not establish that mode selection. Goal-mode operations likewise occur in the retained history. They allowed an explicit objective to remain active across continued work, while completion and blocked states distinguished finished tasks from unresolved requirements. In the modernization record, signing, manual accessibility checks, reference-device measurements, and a performance trade-off remained matters for human or external action. Persistent goals did not remove those limits or authorize the assistant to redefine success (LoopCAT project, 2026; OpenAI, n.d.-b).

Skills and delegated work. Skills provide reusable instructions and supporting resources for a task; they do not retrain the model (OpenAI, n.d.-a). Development records include reads of skills for browser control, Catalan localization, and development coaching during software work. Subagents were used for bounded investigations: the 25 August shortcut-repair task records separate agents for shortcut diagnosis and release-check updates. These were AI-generated contributions requiring reconciliation, not independent human reviews (OpenAI, n.d.-c). Later, preparation of this article used research-review and document-production skills and delegated code, citation, and UI-localization audits. Those manuscript activities are recorded separately from the earlier software development.

Correction through use. The keyboard work shows why specification and verification both needed human input. On 25 August, the developer first requested a shortcut strategy without implementation, then authorized the changes. After the initial checks, the developer reported that Chrome intercepted shortcuts. The subsequent task revised bindings, refreshed the offline cache and web distribution, and added checks against the packaged workflow. Commit d406a20 records the correction. This episode supports a specific account of iteration: a planned and tested change still required correction after use. It does not establish that the AI contribution was uniformly reliable or that automated tests replaced the developer's judgement.

An instructor could adapt a small, sanitized development episode into an exercise: write acceptance criteria, inspect an AI-proposed change, test a counterexample, and justify acceptance, revision, or rejection. Students could contribute through a specification, interface translation, or reproducible defect report without writing code. This extends the paper's evaluative-judgement approach to software creation, but its educational effectiveness remains untested. It also exposes a dependency worth discussing: a local-first, open-source artifact can be developed with a commercial AI service. Codex is part of LoopCAT's development history, not a required service for ordinary use of the application. Human authors and maintainers remain responsible for the claims and changes they accept.

## 5. Architecture and the current evidence boundary

### 5.1. A translation workflow with optional external services

LoopCAT uses HTML, CSS, and JavaScript for a browser interface and an Electron desktop package. Projects, segments, translation-memory entries, terms, and selected activity records are stored in IndexedDB. Where directory access is supported and permission is granted, users can also connect a workspace folder and save portable project packages. The workspace brings together bilingual editing, terminology, quality checks, review, revision and recovery functions, and import/export pathways. The bundled UI languages are English (en-US), Catalan (ca-ES), and Turkish (tr-TR). English source export and custom catalog import allow users to revise the interface language without rebuilding the application; Section 6.3 develops this as a teaching activity. Catalog availability does not establish complete translation or linguistic review.

The ordinary editing workflow need not depend on a hosted AI service. AI connections form an optional part of the environment. Local LLM serving and local NMT have different setup requirements, while hosted providers introduce an additional network and policy boundary. An educator must distinguish the LoopCAT application, the provider application or endpoint, the selected model, and the rights associated with its weights and data. The application's Apache 2.0 license does not make every connected component free or open source.

This organization offers a useful teaching model with three interacting layers: translation work (segments, resources, quality and delivery), local application state (projects, settings, revisions and exports), and optional services (MT or LLM endpoints). The model is explanatory rather than a claim that every implementation module maps neatly to one layer. It encourages students to ask which layer is responsible when a suggestion is inappropriate, a file cannot be recovered, or a request fails.

### 5.2. Human control requires visible consequences

Users need to understand what an action changes and how to revise or recover their work. LoopCAT provides deliberate candidate insertion, review and correction, undo/redo, bounded target histories, recovery paths, and export checks. These mechanisms have different scopes: undo/redo is held in memory, while target history retains up to 25 entries per segment and can combine nearby typing changes. Neither is an exhaustive record of the translation process. QA warnings likewise identify specified conditions; they do not assess every aspect of adequacy, register, or communicative purpose.

Export behaviour illustrates why these distinctions matter. Delivery checks block specified structural and safety risks, while a review export can preserve problems for inspection. The treatment of untranslated content also depends on the pathway. A monolingual delivery export may, after confirmation, use source-text fallback; bilingual interchange can preserve an empty target. Non-empty but unconfirmed targets can be exported as written after confirmation. Students should learn what each output contains rather than assume that the software prevents every incomplete delivery.

Local storage introduces responsibilities concerning backups, shared devices, and recovery. Browser data can be removed and disks can fail; a project in IndexedDB is not an independent backup. The application requests persistent storage where available and warns when storage remains best-effort, but users still need exported copies. The AI configuration needs separate scrutiny: a server reached through a localhost address may forward requests to a cloud model, and requests can include nearby context, memory matches, and terminology hints as well as source text. Local-first design is not a claim of encryption at rest or regulatory compliance. A classroom data-flow exercise should examine the configured route and the information it carries.

### 5.3. What has been verified, and what has not

Two verification records must be distinguished. The first 31 August build, 0.0.4-dev.20260831+source.92e3832c0146, has a record preserved at commit 742467cf54ad4b97a8d93771bb91132c64b98914. It reports 1,389 passing unit tests, eight successful browser-test phases, and 36 passing automated accessibility audits across light and dark themes. The browser phases covered security policy, offline shell, smoke and regression checks, application workflow, workspace storage, package round trips, and a large-project scenario. Web and Windows package identities and contents were verified. An installed-desktop upgrade was also checked on the same Windows machine; the record reports 109 checked storage files unchanged. This is evidence of that local upgrade, not clean-machine qualification.

For the current privacy rebuild, 0.0.4-dev.20260831+source.239cf868ce3d, the fresh checks cover release and build contracts, three focused build-identity tests, archive contents and checksums, rendered web smoke, Electron payload integrity and security fuses, and packaged desktop startup. Startup used isolated temporary profiles, the renderer OS sandbox, and both normal graphics acceleration and a fallback mode. Installer and portable payloads matched the verified unpacked application. The record also reports that 37 renderer/runtime files are byte-identical to the preceding build. The full unit, browser, and accessibility suites and the installed-desktop upgrade were not repeated for this refresh; their earlier results are supporting evidence for the unchanged runtime, not new test runs on the rebuilt packages.

These checks do not establish universal correctness, production safety, or educational effectiveness. Both Windows executables remain unsigned. Clean-machine installation and upgrade testing, broader device and sustained-performance qualification, manual screen-reader evaluation, disk-full and permission-denied scenarios, and native macOS/Linux package qualification remain outstanding. A browser build does not justify treating every operating system and assistive-technology combination as verified.

LoopCAT provides Project Report and Quality Passport exports containing selected QA, resource, and activity information. These may support a portfolio, but the Quality Passport's confidence indicator is a hand-weighted heuristic, not a validated measure of translation quality or learning. The application does not provide a complete classroom research system, native model training, an integrated COMET/BLEU/chrF evaluation suite, or real-time collaborative editing. Teachers can use external tools, such as the MATEO evaluation interface described by Vanroy et al. (2023), if they document the additional setup and data transfers. Exchanging project files between students is also possible, but differs from synchronous collaboration.

## 6. A pedagogical framework: operate, evaluate, inspect, intervene

### 6.1. Four connected forms of participation

The proposed framework organizes activities into four connected forms of participation. They are not fixed developmental stages, nor does success at one guarantee success at another. Students may move between them as a problem requires. An experienced translator can exercise substantial agency without changing code, while a successful code edit can coexist with weak translation judgement.

Operating a workflow means completing a translation task while preserving its resources and delivery constraints. Students should be able to explain the roles of a project, translation memory, termbase, segment status, and export. Evidence includes a reopenable project and an inspected deliverable. Completing a sequence of clicks is insufficient if students cannot recover their work or explain what was delivered.

Evaluating outputs and decisions means establishing criteria, comparing candidates, and justifying acceptance, revision, or rejection. Candidates may come from the student's own translation, a memory, an NMT engine, or an LLM. The point is not to establish a permanent ranking of tools from one text. It is to understand what evidence is relevant to this brief and what remains uncertain. This dimension draws on evaluative judgement while situating it within CAT constraints.

Inspecting and configuring mechanisms means following a bounded path from an observed behaviour to its data, settings, or implementation. A student might locate a terminology entry, identify the destination of an AI request, examine a prompt template, or trace the condition that raises a QA warning. Explanation should include a prediction: what would change if this resource or setting changed? A guided trace through one function can be educationally meaningful without requiring comprehensive programming competence.

Intervening in a design means proposing and testing a justified change, or explaining why a change should not be made. The intervention may involve a resource, documentation, a configuration, a reproducible issue report, or a small code patch. Its value depends on the problem it addresses and the evidence for its effects. A new feature is not intrinsically better than a clearer warning, a corrected termbase, or a decision to retain an existing safeguard.

Table 2. From an application feature to assessment evidence.

| Participation | Example learning activity | Evidence and limits |
|---|---|---|
| Operate | Translate, save, reopen, and export a small localization project | Project and deliverable pass a teacher-supplied checklist; no inference of critical understanding from completion alone. |
| Evaluate | Compare human, TM, and optionally AI candidates against a brief | Annotated decisions identify a problem, supporting evidence, and a justified action; no reward for choosing a particular provider. |
| Inspect/configur e | Trace a QA rule or map a provider request and its data | Diagram or explanation predicts one changed condition correctly; source access alone does not demonstrate understanding. |
| Intervene | Revise a UI catalog, resource, documentation item, or bounded feature | Before/after example, acceptance criteria, and regression check; no requirement to publish code or use a paid assistant. |

Table 2 is the article's proposed teaching framework, informed by Tai et al. (2018), Bearman et al. (2024), and Hackenbuchner and Krüger (2023). It has not been validated as an assessment instrument.

### 6.2. A six-session sequence

The sequence below is designed as a configurable module for a translation technology course. A feasible starting allocation is six sessions of approximately two hours, with preparation and portfolio work between meetings; this is a planning assumption, not an observed teaching duration. The instructor should reduce text length and technical depth where students have limited CAT experience. A core route requires no AI subscription or local model, while an optional route introduces provider comparison. Both routes should carry equivalent credit.

Session 1: Establish and recover a workflow. Students receive a short, licensed or synthetic source file, a translation brief, and a small resource pack. They identify the relationship between the source, project, memory, termbase, and final deliverable. After initial editing, they close and reopen the project and inspect an export. The portfolio item is a workflow map and a recovery note explaining which files or backups are needed. The instructor checks setup barriers before assessing speed or independence.

Session 2: Compare evidence for translation decisions. Students compare their own solutions with supplied candidates, including deliberately imperfect memory entries. Where permitted, they add NMT or LLM output and record the provider, model, prompt, and date. They identify semantic, terminological, stylistic, and structural issues, then explain why a candidate should be retained, revised, or rejected. Peers compare rationales using the same brief. Automatic metrics, if used, are obtained in a separately documented external exercise and discussed as additional evidence rather than authoritative grades.

Session 3: Investigate a quality rule. A teacher-supplied example triggers a placeholder or terminology warning. Students distinguish the condition detected by the rule from the wider translation problem. With a prepared source-code map, they locate the relevant condition and test a contrasting example. Their portfolio explains one true positive, one possible false alarm or uncovered case, and the limits of the rule. The aim is to understand an operational definition of quality, not to defeat a safeguard so that an export succeeds.

Session 4: Map data and choose an AI configuration. Students trace a source segment and any accompanying resources through the application and provider. They check whether a local server forwards requests to a cloud model and whether nearby text, memory matches, or terminology hints are included. A simulated confidentiality constraint requires them to justify a local, hosted, or no-AI route. The instructor supplies the policy conditions. A valid submission identifies uncertainty, such as incomplete provider documentation, and states what would need checking before real data could be used.

Session 5: Propose and test a bounded improvement. Students select a problem from earlier sessions or localize a small part of LoopCAT's own interface (Section 6.3). Possible contributions include a revised UI catalog, a clearer warning, recovery documentation, a terminology correction, or a small QA patch in an isolated copy. Students explain the change and demonstrate that relevant existing behaviour still works. A coding assistant may help with a patch, but an unexplained patch is insufficient evidence. A specification and reproducible issue report provide an equivalent non-code route.

Session 6: Demonstrate transfer and defend a decision. Students exchange sanitized project packages or inspect equivalent material in another available CAT environment. They identify which concepts transfer and which behaviours differ. Each student defends one translation decision and one infrastructure decision, then revises a portfolio explanation after peer feedback. The resulting evidence concerns reasoning and transfer, not loyalty to LoopCAT or the number of features used.

### 6.3. Localizing the tool used for learning

LoopCAT can itself become a small software-localization project. The English source is available as i18n/source.en-US.json and through Workspace → Interface language → Export UI source. It includes stable message identifiers, message text, descriptions, and source locations. Import UI translation loads and selects a custom JSON catalog in the running application without rebuilding it. Students can therefore work on text they will subsequently encounter while using the tool.

Two assignments are possible: translating a selected group of English messages into another language, or post-editing the existing Catalan or Turkish catalog. The developer reports that the latter were generated automatically; the files themselves do not record the generating model, prompts, or review history. They should be treated as material for assessment, not as reference translations. The inspected target catalogs contain empty entries and mixed-language wording, giving students concrete problems to investigate. Students who do not know Catalan or Turkish can translate from English into another language they know.

Prepare a small, contextualized task. The instructor selects messages from one workflow, such as project creation or document import, and supplies a brief describing users, register, terminology, and the screens to test. Students inspect those screens before translating. A description or source location can help establish context, but some locations are historical and do not replace checking the running interface. For post-editing, the instructor supplies a copy of the relevant file in

i18n/locales/ and asks students to justify selected changes against English and the actual function.

Translate and import a classroom copy. Students preserve message identifiers, JSON syntax, and named placeholders while revising message values. Work should use a dedicated profile containing only teaching data and a distinct locale identifier, for example ca-ES-x-class. This avoids replacing a bundled language in that profile. Reimporting the same locale replaces its whole catalog rather than merging incremental changes, so students should maintain a complete classroom copy outside the profile. Untranslated keys may fall back to another available language; a usable screen does not prove translation completeness.

Test language in use. Students follow the selected workflow and assess action labels, terminology, register, tooltips, text fit, keyboard operation, placeholder values, and singular/plural messages at representative counts. They record defects with reproduction steps and compare the interface before and after revision. Successful import alone establishes little: runtime import does not enforce placeholder matching, and the repository's separate validator permits empty target values. Automated checks, linguistic review, and functional testing therefore answer different questions. The deliverables are a revised catalog, reasons for selected decisions, reproducible defect reports, and a retest record.

Students may edit the catalog in a JSON editor or use LoopCAT's own localization editor. The second route requires teacher preparation. The generic JSON importer extracts string values from descriptions and source-location metadata as well as the UI messages, making the full source export unnecessarily large for a short exercise. A teacher-prepared messages-only subset can be translated in LoopCAT and then wrapped as a catalog containing locale, an optional label, and messages. This is a supported combination of generic JSON translation and custom UI import, not a dedicated self-localization wizard. Packaging a new language as a built-in option is a separate development task requiring registration, rebuilding, and testing.

Interface translation and testing are established contribution routes in open CAT projects such as OmegaT (OmegaT project, n.d.). Here they provide a concrete non-code form of intervention: students can improve the environment in which they work and explain how a wording choice affects an action. The activity is proposed for teaching; the module-level checks performed for this paper establish technical feasibility, not successful classroom deployment.

### 6.4. Worked example: a fluent target with a missing placeholder

Consider a synthetic software-localization task with the English source string **Delete {count} files?**. The brief requires **{count}** to remain unchanged and specifies an approved Catalan term for the action. A supplied candidate, **Vols eliminar els fitxers seleccionats?**, sounds plausible but omits the numerical placeholder. This example is invented for teaching, not attributed to a model or students. An isolated check of the actual LoopCAT detector and QA function recognized **{count}** and flagged its omission. Instructors should still test the chosen import/export route before class. The problem also has an empirical precedent: Dogru and Moorkens (2024, Section 4.1, Table 9) discuss damaged placeholders in English–Catalan localization output from NMT systems. That study did not evaluate LoopCAT's interface translations.

At the operational level, the student imports the material, edits the target, runs the relevant checks, and inspects the exported string. At the evaluative level, the student distinguishes fluency from compliance with the brief: an acceptable-looking sentence has lost information required by the interface. A possible corrected candidate is **Vols eliminar {count} fitxers?**, subject to the actual product's register, terminology, and pluralization rules. This is not offered as a complete solution to ICU plural handling or every count value.

At the inspection level, the student traces how the application recognizes the protected element, compares source and target, and presents a warning or export constraint. The question becomes: what does the rule know, and what does it not know? Preserving the placeholder cannot guarantee an appropriate register, correct terminology, or the intended consequences of the button. Conversely, a structurally safe export is not proof of an adequate translation.

At the intervention level, the student might improve the explanatory text associated with the warning, add a documented counterexample, or propose a regression case for a supported placeholder form. A code route requires review in a disposable copy; a non-code route specifies the defect and expected behaviour. In either route, assessment concerns whether the student identifies the cause, proposes a defensible response, and checks its consequences.

### 6.5. Assessing decisions and their justification

Assessment should combine the final artifact with an explanation of selected decisions. A portfolio can contain the project and export, an annotated candidate comparison, a data-flow or rule explanation, and an intervention with evidence. For UI localization, the catalog, defect reports, and retest record supply the latter evidence. Project Reports and Quality Passports can contribute workflow information, but do not replace students' rationales or the teacher's assessment. The instructor should state what students must preserve and offer accessible alternatives to screenshots or video.

Table 3. Proposed analytic rubric for a selected decision.

| Dimension | Limited evidence | Developing evidence | Strong evidence |
|---|---|---|---|
| Workflow and delivery | Follows steps but cannot establish what was saved or delivered | Produces usable files and explains the main resources | Verifies delivery constraints and explains recovery or transfer |
| Evaluative judgement | Accepts or rejects a candidate without relevant reasons | Uses explicit criteria and identifies a supported problem | Weighs conflicting evidence, justifies the action, and states uncertainty |
| Mechanism and data understanding | Relies on a tool label or unsupported assumption | Identifies relevant data, settings, or a rule | Traces the mechanism and predicts the effect of a changed condition |
| Responsible intervention | Suggests a change without purpose or verification | Proposes a relevant change with acceptance criteria | Tests the change or justified refusal, considers consequences, and preserves safeguards |

This rubric is a proposed instrument, not a validated scale. “Not observed” should be recorded separately from weak performance; students should not be penalized for an inaccessible optional activity.

Teachers can score the three evidence levels as 1–3 for formative use, provided they explain the anchors and calibrate judgements using shared examples. A single total should not obscure the profile: technically confident students may reason poorly about confidentiality, while careful translators may need support navigating source files. Time, confidence, and number of edits can provide contextual information, but should not stand in for quality of reasoning. Rejecting a proposed change can merit the strongest rating if the student demonstrates that it would damage a necessary safeguard.

## 7. A protocol for future educational evaluation

### 7.1. Research questions and design

A functioning application and feasible activities do not establish learning outcomes. A future study should ask whether students can explain translation and infrastructure decisions more effectively after the module, whether that understanding transfers to another tool or unfamiliar task, and which technical or institutional barriers limit participation. A further question is whether access to implementation details contributes something beyond equivalent instruction using interfaces, documentation, and teacher explanations.

An initial feasibility study could use the six-session module in a course with an explicit pre-task, post-task, and delayed transfer task. The baseline should assess CAT experience, programming familiarity, AI use, language background, and access to suitable hardware. Tasks should use comparable briefs and difficulty while avoiding simple repetition of the same text. The primary outcome should be a preregistered, independently scored measure of decision quality, supported by the rubric after pilot refinement; workflow completion and perceived confidence should be secondary, distinct outcomes.

A stronger comparative study would hold the text, teaching time, criteria, and available candidates as constant as possible while varying access to the inspection/intervention activities. Assignment could occur at an appropriate course-group level where feasible, with attention to contamination between conditions. The comparison should be described as a comparison of teaching arrangements, not automatically of software brands. A small convenience cohort would justify exploratory estimates and qualitative explanation rather than a general claim of effectiveness. Sample size should be justified before recruitment through the intended analysis and available precision, not selected retrospectively to fit an observed effect.

### 7.2. Evidence, analysis, and safeguards

Potential evidence includes sanitized project artifacts, decision rationales, task-based explanations, targeted observations, and interviews. Two assessors should calibrate on pilot work, score independently where practicable, and report agreement as well as how disagreements were handled. Quantitative reporting should include uncertainty and missing-data patterns. Qualitative analysis should actively seek cases that contradict the proposed mechanism: students who understand a rule without reading code, students who make a patch they cannot explain, or students whose reasoning improves despite choosing no AI connection.

Version control is essential to interpretation. The study record should preserve the application build, source fingerprint, operating environment, task materials, resource versions, and any external model or metric configuration. AI outputs vary with model versions, prompts, and settings; a repeatable task must not depend on an undocumented live service. Where licenses and consent allow, static candidates can be released with the teaching pack to support a low-resource replication. Study materials should record when external evaluation or coding tools were used rather than attributing their functions to LoopCAT.

Research participation must be distinguished from course assessment. The teacher–researcher relationship can affect consent and student behaviour, so recruitment and access to identifiable data should be arranged to reduce that pressure. Students should have equivalent learning and grading opportunities without donating research data, creating a public GitHub account, purchasing AI access, or publishing their work. Ethics review and institutional approval should be obtained where required before recruitment; no such approval is claimed for a study that has not yet been conducted.

No participant texts, client material, personal data, or credentials should be placed in public issue reports or sent to external assistants merely to complete an exercise. Synthetic or appropriately licensed materials should be used for public demonstrations. Data retention, access, anonymization, and withdrawal procedures need to be specified in advance. Local storage is one element of this plan, not a substitute for it. Failure or abandonment of a task should be analyzed as a possible access or design problem, rather than automatically interpreted as a student's lack of agency.

## 8. Discussion: opportunities, tensions, and limits

### 8.1. Inspectability has to be taught

The UI-localization exercise shows how a CAT environment can be both a working tool and an object of study. Students start with a wording or workflow problem, examine the relevant message or rule, and test a change. The teaching depends on that connection between use and explanation. Without that connection, students may encounter a repository as another opaque object.

This has implications for documentation and curriculum design. Educators need small, maintained code maps, stable sample projects, clear accounts of where data go, and examples of both successful and unsuccessful interventions. An open-ended "read the source" assignment would privilege prior technical experience. A bounded trace can distribute participation more fairly, particularly when paired with a non-code route. The project's internal restructuring supports this possibility, but does not itself supply a complete teaching package.

### 8.2. Openness can redistribute costs rather than remove them

Free software can remove a license fee and permit adaptation, yet hardware, installation, maintenance, teacher preparation, and support still have costs. Local LLMs may be impractical on some student devices. Institutions may restrict installation, unsigned executables may generate warnings, and accessibility barriers may remain even when the source is available. These are part of the educational design problem, not exceptions to be ignored.

The UNESCO Recommendation on Open Educational Resources emphasizes access, reuse, adaptation, and redistribution alongside capacity building and sustainable provision (UNESCO, 2019). An Apache-licensed application is not, by itself, an openly licensed curriculum. Teaching packs require their own licensing and accessibility decisions, including permissions for example texts and images. The relationship with Sustainable Development Goal 4 is therefore an orientation toward inclusive educational provision, not a demonstrated impact. Actual access should be evaluated through participation barriers, support needs, and the availability of equivalent learning routes.

### 8.3. Agency remains relational and constrained

Students' choices occur within briefs, assessment systems, institutional policy, and professional power relationships. A translator who understands a cloud service may still be required to use it. A student who proposes a useful feature may lack the authority to deploy it. The framework should therefore include the ability to document a concern, communicate a requirement, and negotiate a condition, not only the technical ability to change a setting or file.

The co-creation process adds a related responsibility: an AI-generated patch and an AI-generated test may share the same mistaken assumption. A passing test therefore needs to be considered alongside the brief, code inspection, counterexamples, and testing in use. Open source also introduces responsibilities. A fork may become difficult to maintain; a seemingly helpful patch may break import/export behaviour; an AI-generated change may create a security or privacy problem. Teaching intervention should include scope, review, regression checking, and an explanation of what has not been verified. This is consistent with treating coding assistants as tools whose output requires judgement, rather than as a shortcut around understanding.

### 8.4. A platform claim must remain proportionate to the evidence

The present account cannot establish comparative usability, sustained institutional adoption, improved employment outcomes, or superior translation quality. Its documentary method is vulnerable to selective reconstruction, and the proposed rubric requires validation. Engineering tests provide valuable evidence for a specific snapshot, but do not measure educational effectiveness. The absence of a LoopCAT classroom study is a substantive limitation, not something that can be filled by citing a study conducted with different tools.

The paper provides activities and an evaluation plan that other educators can inspect and challenge. The framework would be weakened if students could complete inspection tasks without understanding consequences, if technical setup systematically excluded participants, or if comparable gains arose from simpler activities requiring no source access. Reporting these possibilities makes it possible to investigate when LoopCAT is useful, for whom, and at what cost.

## 9. Conclusion

LoopCAT's development began, in the available record, with the practical objective of sustaining a complete local translation workflow. Subsequent work broadened its resources, AI connections, interface, internal organization, and release verification. This software was co-created with OpenAI Codex through planning, implementation, testing, and correction directed by the developer. These developments create opportunities to teach translation technology through decisions that students can observe, explain, and sometimes change.

The article has proposed a framework connecting workflow operation, evaluative judgement, inspection and configuration, and responsible intervention. The sequence and rubric specify activities and assessment evidence, including translation, post-editing, and functional testing of the English, Catalan, and Turkish interface resources. They preserve non-code and no-AI routes.

They also make clear what remains to be established: educational effectiveness, transfer, accessibility, and sustainable classroom use require evidence from future research.

The educational ambition is not to make every translator a software developer. It is to help future language professionals understand the systems they work with, justify their use or refusal, communicate better requirements, and contribute informed changes when circumstances permit. A local-first, open-source CAT platform can support that ambition when inspectability is accompanied by dependable workflows, appropriate scaffolding, and critical attention to the limits of individual control.

## Declarations and availability

Software and materials. LoopCAT is licensed under Apache 2.0. Its canonical repository is https://github.com/gokhandogru/LoopCAT. The current development preview is 0.0.4-dev.20260831+source.239cf868ce3d, with source and download-record commits identified in Section 3. An authenticated GitHub check on 31 August confirmed that the repository remains private.

AI-assisted software development. LoopCAT was co-created by the authors with OpenAI Codex using GPT-5.5 and GPT-5.6 models, including the recorded GPT-5.6 Sol identifier, with varying reasoning-effort settings and developer-reported speed choices. Codex made substantive contributions to design, code, refactoring, tests, and documentation. Section 4.4 describes examples supported by task records and distinguishes those records from the developer's account. Responsibility for the accepted software and its release remains with its human maintainers.

AI-assisted manuscript preparation. Codex also helped inspect project materials, discover and check sources, draft and revise this article, audit its claims, and produce the document, using task-specific skills and subagents.